\documentclass[10pt,twocolumn,letterpaper]{article}

\usepackage{cvpr}
\usepackage{times}
\usepackage{epsfig}
\usepackage{graphicx}
\usepackage{amsmath}
\usepackage{amssymb}

\usepackage{bm}

\usepackage{color}
\usepackage{array}
\usepackage{comment}
\usepackage{cases}
\usepackage{multirow}
\usepackage{textcomp}
\usepackage{subfigure}
\usepackage[linesnumbered, ruled]{algorithm2e}
\usepackage{algpseudocode}
\usepackage{booktabs}
\usepackage{colortbl}
\usepackage{epstopdf}

\def  \Isa	   {I_i}						 
\def  \Rsa	   {R_j}					 	 
\def  \Ssa	   {S_k}					 	 
\def  \Rgt	   {R(I_i)^{GT}}				 
\def  \Sgt	   {S(I_i)^{GT}}				 
\def  \Resm	   {R(I_i)}     	     	     
\def  \Sesm	   {S(I_i)}	    		         
\def  \Idm     {\mathcal{I}}                 
\def  \Rdm     {\mathcal{R}}                 
\def  \Sdm     {\mathcal{S}}                 
\def  \Cdm     {\mathcal{C}}                 
\def  \Rpdm    {\mathcal{Z}_\mathcal{R}}     
\def  \Spdm    {\mathcal{Z}_\mathcal{S}}     
\def  \GenI    {G_\mathcal{I}}               
\def  \GenR    {G_\mathcal{R}}               
\def  \GenS    {G_\mathcal{S}}               
\def  \EnIp    {E_\mathcal{I}^p}             
\def  \EnIc    {E_\mathcal{I}^c}             
\def  \EnRp    {E_\mathcal{R}^p}             
\def  \EnRc    {E_\mathcal{R}^c}             
\def  \EnSp    {E_\mathcal{S}^p}             
\def  \EnSc    {E_\mathcal{S}^c}             
\def  \zR      {z_{R_j}}			         
\def  \zS      {z_{S_k}}			         
\def  \E       {\mathbb{E}}		             
\def  \LL      {\mathcal{L}}                 
\def  \xx      {\textbf{x}}                  

\usepackage{color}

\usepackage{lipsum}

\definecolor{mygray}{gray}{.9}
\definecolor{mypink}{rgb}{.99,.91,.95}
\definecolor{mycyan}{cmyk}{.3,0,0,0}

\newcommand{\tabincell}[2]{\begin{tabular}{@{}#1@{}}#2\end{tabular}}

\newcommand{\yf}[1]{{\textcolor[rgb]{0.0,0.0,0.0}{#1}}}

\newcommand{\ours}[1]{\multicolumn{1}{>{\columncolor{mygray}}c}{{\bf #1}}}
\newcommand{\proposed}{USI$^3$D}

\usepackage[pagebackref=true,breaklinks=true,letterpaper=true,colorlinks,bookmarks=false]{hyperref}

\cvprfinalcopy 

\def\cvprPaperID{2640} 

\ifcvprfinal\fi
\begin{document}
	
	\title{Unsupervised Learning for Intrinsic Image Decomposition from a Single Image}
	
	\author{Yunfei~Liu\textsuperscript{\rm 1}\qquad~Yu~Li\textsuperscript{\rm 2}\qquad~Shaodi~You\textsuperscript{\rm 3}\qquad~Feng~Lu\textsuperscript{\rm 1, 4, }\thanks{ Corresponding Author. 
			\newline \indent~ This work is partially supported by the National Natural Science Foundation of China (NSFC) under Grant 61972012 and Grant 61732016, and Baidu academic collaboration program.
		    }\\
		{\textsuperscript{\rm 1} State Key Laboratory of VR Technology and Systems, 
		School of CSE, Beihang University}  \\
		{\textsuperscript{\rm 2}Applied Research Center (ARC), Tencent PCG} ~~
		{\textsuperscript{\rm 3}University of Amsterdam, Amsterdam, Netherland} \\
		{\textsuperscript{\rm 4}Peng Cheng Laboratory, Shenzhen, China} \\
		\small{\texttt{\{lyunfei,lufeng\}@buaa.edu.cn}} \qquad
		\small{\texttt{ianyli@tencent.com}} \qquad
		\small{\texttt{s.you@uva.nl}}
	}
	
	\maketitle
	
	\begin{abstract}
		Intrinsic image decomposition, which is an essential task in computer vision, aims to infer the reflectance and shading of the scene. It is challenging since it needs to separate one image into two components. To tackle this, conventional methods introduce various priors to constrain the solution, yet with limited performance. Meanwhile, the problem is typically solved by supervised learning methods, which is actually not an ideal solution since obtaining ground truth reflectance and shading for massive general natural scenes is challenging and even impossible.
        In this paper, we propose a novel unsupervised intrinsic image decomposition framework, which relies on neither labelled training data nor hand-crafted priors. Instead, it directly learns the latent feature of reflectance and shading from unsupervised and uncorrelated data. To enable this, we explore the independence between reflectance and shading, the domain invariant content constraint and the physical constraint.
        Extensive experiments on both synthetic and real image datasets demonstrate consistently superior performance of the proposed method.
	\end{abstract}

	
	\section{Introduction}
	The appearance of a natural image depends on various factors, such as illumination, shape and material. 
	Intrinsic image decomposition aims to decompose such a natural image into an illumination-invariant component and an illumination-variant component.
	Therefore, it can benefit a variety of high-level computer vision tasks such as texture editing~\cite{PPP:Bi2015IntrinsicDecompositionTOG}, face appearance editing~\cite{PPP:Cao2017symps} and many others.  
	In this paper, we follow the common practice~\cite{PPP:Fan2018revisitingDeepIntrinsic,PPP:Li2014Single,PPP:narihira2015direct_MSCR} that assumes the ideal Lambertian surface. Then, a natural image $I$ can be decomposed as the pixel-wise product of the illumination invariance, the reflectance $R(I)$; and the illumination variance, the shading $S(I)$, \ie.,
	\begin{equation} \label{eq:intro}
	I = R(I) \odot S(I).
	\end{equation}
	
	\begin{figure}[!t]
		\begin{center}
			\includegraphics[width=\linewidth]{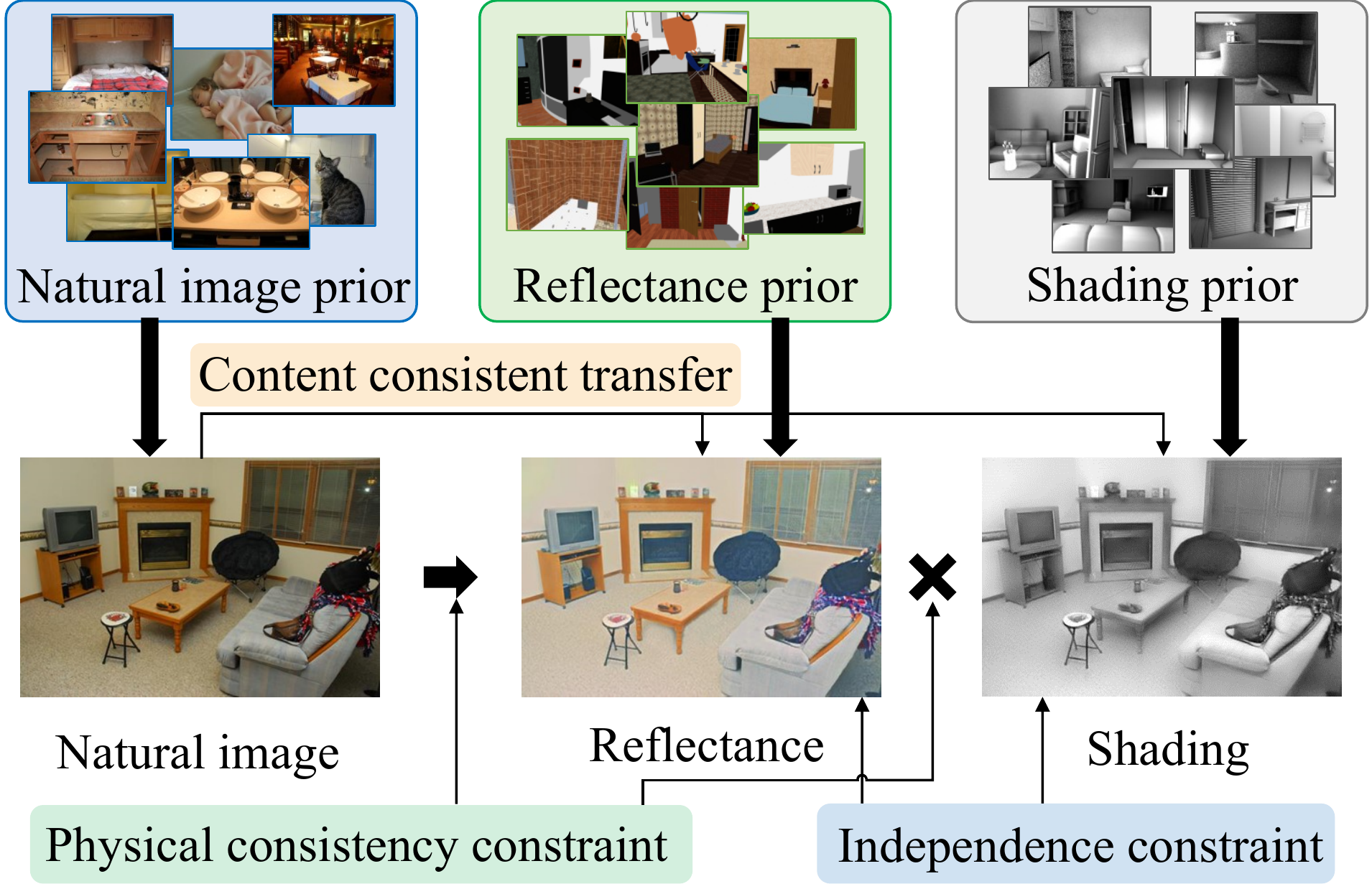}
		\end{center}
		\caption{Our method learns intrinsic image decomposition in an unsupervised fashion where the ground truth reflectance and shading is not available in the training data. We learn the distribution priors from unlabeled and uncorrelated collections of natural image, reflectance and shading. Then we perform intrinsic image decomposition through content preserving image translation with independence constraint and physical consistency constraint.}
		\label{fig:1-teaser}
		\vspace{-3mm}
	\end{figure}
	
	Eq.~\eqref{eq:intro} is ill-posed because there are twice the unknowns than the knowns. 
	Conventional methods ~\cite{PPP:Bi2015IntrinsicDecompositionTOG,PPP:Li2014Single} therefore explore physical priors as extra constraints, while recent researches tend to use deep neutral networks to directly learn such priors \cite{PPP:Fan2018revisitingDeepIntrinsic,PPP:narihira2015direct_MSCR,PPP:Shi2017learningNonLambertian,PPP:Zhou2015dataDrivenIntrinsic}.
	
	Unlike high-level vision tasks, intrinsic image decomposition is obviously physics-based, and therefore designing a supervised learning method will heavily rely on high quality physical realistic ground truth. However, existing datasets are either created from a small set of manually painted objects ~\cite{PPP:Grosse2011MITintrinsic}, synthetic objects or scenes~\cite{PPP:Butler2012MPI_Sintel,PPP:Su2015ShapeNet,PPP:li2018cgintrinsics} or manual annotations~\cite{PPP:kovacs17SAW,PPP:bell14IIW}. These datasets are either too small or far from natural images and therefore limit the performance of supervised learning.
	
	A few semi-supervised and unsupervised learning methods have been exploited very recently.
	Janner \etal~\cite{PPP:Michael2017SelfSupervisedIntrinsit} proposed self-supervised intrinsic image decomposition which relies on few labelled training data and then transfers to other unlabelled data. However, many other supervised information such as shape of the object need be involved.
	Li \etal \cite{PPP:Li2018BigTime} and Ma \etal \cite{PPP:Ma2018IntrinsicDecompositionWOsingle} 
	work on unlabelled image sequences where the scene requires to be fixed within the sequence and only the lighting and shading allow to change. Such settings are still very limited.
	
	In this paper, we aim to explore single image unsupervised intrinsic image decomposition.
	The key idea is that the natural image, the reflectance and the shading all share the same content, which reflects the nature of the target object in the scene. Therefore, we consider estimating reflectance and shading from a natural image as transferring image style but remaining the image content.
	Based on such an idea, we can actually use unsupervised learning method to learn the style of natural image, reflectance and shading by collecting three unlabelled and un-correlated samples for each set. Then we apply auto-encoder and generative-adversarial network to transfer the natural image to the desired style while preserve the underlying content.
	Unlike na\i"ve unsupervised style transfer methods which are from one domain to another, our method transfers from one domain to another two domains with explicit physical meanings. We therefore explicitly adopt three physical constraints into our proposed method which are 1) the physical consistent constraint as in Eq.~\eqref{eq:intro}, 2) domain invariant content constraint that natural image and its decomposition layers share the same object, layout, and geometry, 3) the physical independent constraint that reflectance is illumination-invariant and shading is illumination-variant.
	
	Rigorous experiments show that our method can produce superior performance against state-of-the-art unsupervised methods on four benchmarks, namely ShapeNet, MPI Sintel benchmark, MIT intrinsic dataset and IIW. Our method also demonstrates comparable performance with state-of-the-art fully supervised methods~\cite{PPP:Fan2018revisitingDeepIntrinsic,PPP:Shi2017learningNonLambertian}, and even outperforms some of them appeared in the recent years~\cite{PPP:narihira2015direct_MSCR,PPP:Zhou2015dataDrivenIntrinsic}.
	
	The contributions of this work are threefold:
	\begin{itemize}
		\item To the best of our knowledge, we propose the first physics based \textit{single image} unsupervised learning for intrinsic image decomposition. 
		Specifically, we adopt three physical constraints: the physical consistency constraint, the domain invariant content constraint and the reflectance-shading independence. 
		\item We propose and implement a completely unsupervised learning network architecture for single image intrinsic decomposition.
		\item The proposed method outperforms existing unsupervised methods and shows comparable results against fully supervised methods on different intrinsic image benchmarks.
	\end{itemize}

	
	\section{Related Works}
	
	\noindent\textbf{Optimization based methods.}
	Intrinsic image decomposition is \yf{ a typical image layer separation problem~\cite{PPP:Land1971retinex, PPP:Li2014Single,PPP:Liu2019SGRRN,PPP:Liu2020SILS}} which
	has been studied for nearly fifty years.
	To handle the ill-posed problem, additional priors with an optimization framework have been applied.
	For instance, Land \etal~\cite{PPP:Land1971retinex} propose a seminal Retinex algorithm which assumes large image gradients corresponding to changes in reflectance, while smaller gradients are with shading. 
	Subsequently, many priors for intrinsic image decomposition have been explored. Inspired by the large image gradients and piece-wise constant property, reflectance sparsity~\cite{PPP:Rother2011RecoveringIntrinsic,PPP:Shen2011IntrinsicImages} and low-rank reflectance~\cite{PPP:Bousseau2015User_assisted} are taken as regularization term in object functions.
	There are also many constraints for shading such as the distribution difference in gradient domain~\cite{PPP:Bi2015IntrinsicDecompositionTOG,PPP:Li2014Single}.
	Recently, Chen \etal~\cite{PPP:Cheng_2019_NearInfrared} use near-infrared image and propose near infrared prior to regularize the intrinsic image decomposition.
	Although these hand-crafted priors are reasonable in small image set, they are not likely to cover complex scenes~\cite{PPP:kovacs17SAW,PPP:bell14IIW}.
	\yf{Furthermore, the above mentioned methods assume Lambertian material and thus cannot be adopted to more complex situations with general non-Lambertian reflectance~\cite{PPP:Lu2017symps,PPP:Lu2015intensity}.}
	
	\noindent\textbf{Supervised learning methods.}
	Many supervised learning based methods have been proposed in recent years,
	\cite{PPP:Fan2018revisitingDeepIntrinsic,PPP:li2018cgintrinsics,PPP:narihira2015direct_MSCR,PPP:Shi2017learningNonLambertian,PPP:Wang2019Intrinsic,PPP:Zhou2015dataDrivenIntrinsic}. These methods try to estimate the reflectance and shading on labelled training data with different network architectures.
	However, the training data from publicly available datasets have obvious shortcomings: Sintel~\cite{PPP:Butler2012MPI_Sintel}, ShapeNet~\cite{PPP:Su2015ShapeNet} and CGIntrinsics~\cite{PPP:li2018cgintrinsics} are highly synthetic datasets, and networks trained on them cannot generalize well to real-world scenes. The MIT intrinsic dataset~\cite{PPP:Grosse2011MITintrinsic} consists of real images, but the number of these images are too limited since it contains just 20 objects with ground truth.
	As a result, these supervised methods often cannot generalize well on one dataset if they are trained on another dataset.
	Recently, human-labelled dataset IIW~\cite{PPP:bell14IIW} and SAW~\cite{PPP:kovacs17SAW} only contain sparse annotations. Not only that, it is difficult to collect such annotations at scale.
	
	\noindent\textbf{Semi-supervised and unsupervised learning methods.}
	Involving other supervised information such as shape, illumination source of the object, Janner \etal~\cite{PPP:Michael2017SelfSupervisedIntrinsit} propose self-supervised intrinsic image decomposition, which relied on few labeled training data and then transferred to other un-labeled data.
	InverseRenderNet~\cite{PPP:yu2019inverserendernet} uses a series of related images as input and adds multi-view stereo as supervised signal to intrinsic decomposition.
	Leveraging videos or image sequences, together with physical constraints, learning without intrinsic image has recently become an emerging topic of research. 
	The most of existing unsupervised intrinsic image decomposition methods~\cite{PPP:Li2018BigTime,PPP:Ma2018IntrinsicDecompositionWOsingle} are mainly focus on training on images with fixed scene and varied illumination, then the trained model can be tested with single input.
	However, the scene-level real images for training are still limited to various scenes in most cases.
	Motivated by this, we propose an alternative unsupervised method, which doesn't rely on fixed structure image sequences to train.
	
	\noindent\textbf{Image-to-image translation.}
	Image-to-image translation aims to learn the mapping from two image domains.
	Pix2pix~\cite{UII:Isola2017imageCGAN} employs conditional GAN to learn the mapping, CycleGAN~\cite{UII:hu2017unpairedCycleGAN}, UNIT~\cite{UII:Liu2017UnsupervisedImageTranslation} applies the cycle consistency to regularize the training.
	More recently, MUNIT~\cite{UII:Huang2018MUNIT} and DRIT~\cite{UII:Lee2018Diverse} assume \emph{partially shared latent space assumption} and make multi-modal image-to-image translation.
	However, there is still a great gap between unsupervised image-to-image translation between and intrinsic image decomposition because 1) image-to-image are fully statistics driven whereas intrinsic image decomposition is physic-based, 2) the translated image can be various of modalities while the intrinsic
	images of an input image are explicit. Thus, the image-to-image translation method is not directly adaptable to intrinsic image decomposition.

	
	\section{Unsupervised Single Input Intrinsic Image Decomposition}
	
	\subsection{Problem formulation and assumptions}
	
	\paragraph{Single input intrinsic image decomposition.}
	To begin with, we formulate the task with precise denotations.
	As illustrated in Fig.~\ref{fig:1-teaser} and Eq.~\eqref{eq:intro}, the goal of single image intrinsic decomposition is to decompose a natural image, denoted as $I$, into two layers, illumination-invariance, namely the reflectance $R(I)$; 
	and illumination-variance, namely the shading $S(I)$.
	
	Eq.~\eqref{eq:intro} has more `unknowns' than `knowns' and therefore is not directly solvable.
	Providing sufficient amount of data-samples with ground truths, \ie, the triplet samples $\{(\Isa, \Rgt, \Sgt)\}$, supervised learning based methods have also been explored\cite{PPP:Fan2018revisitingDeepIntrinsic,PPP:li2018cgintrinsics,PPP:narihira2015direct_MSCR,PPP:Shi2017learningNonLambertian}. 
	In previous sections, we have discussed the difficulty in obtaining the ground truth. We now focus on unsupervised learning. 
	
	\begin{figure} [!tb]
		\centering
		\includegraphics[width=0.6\linewidth]{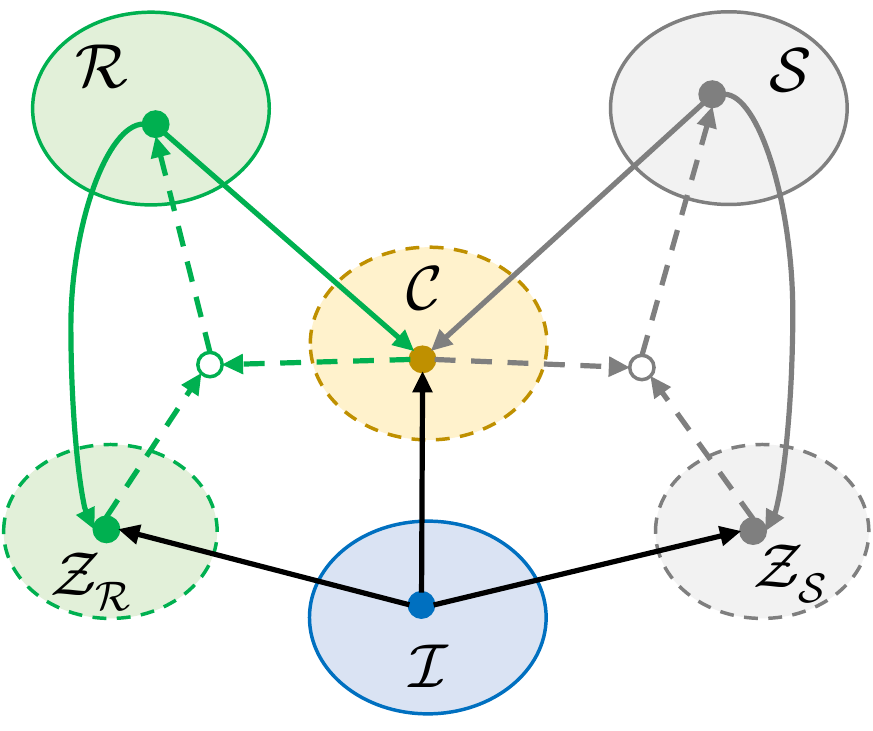}
		\caption{Content preserving translation among domains. $\mathcal{I}$ is the domain of natural image. $\mathcal{S}$ is the domain of shading and $\mathcal{R}$ is the domain of reflectance. 
			For our unsupervised learning method, we learn a set of encoders which encode the appearance from each domain to domain-invariance latent space $\mathcal{C}$. We also learn the encoders to encode the appearance to domain-depended prior space for reflectance ($\mathcal{Z_R}$) and shading ($\mathcal{Z_S}$) correspondingly. Later the image style can be transferred from encoders (solid arrows) to generators (dash arrows).
		}
		\label{fig:concept}
		\vspace{-3mm}
	\end{figure}
	
	\paragraph{Unsupervised Intrinsic Image Decomposition.}

	\yf{In this section, we define the Unsupervised Single Image Intrinsic Image Decomposition (\proposed) problem.} Assuming we collect unlabelled and unrelated samples, we learn the appearance style of each collection. Say, we can learn the style of reflectance, the marginal distribution $p(\Rsa)$, by providing a set of unlabelled reflectance images: $\{\Rsa \in \Rdm\}$; we learn the shading style, the marginal distribution $p(\Ssa)$, by providing a set of unlabelled shading images: $\{\Ssa \in \Sdm\}$; and we learn the natural image style, marginal distribution $p(\Isa)$, by providing a set of unlabelled natural images: $\{\Isa \in \Idm\}$.
	Then, we aim to infer $\Resm$, $\Sesm$ of $I_i$ from the marginal distributions.

	To make the task tractable, we make the following three assumptions.

	\noindent\textbf{Assumption-1. Domain invariant content.} \\ Physically, the natural appearance, the reflectance and the shading are all the appearance of a given object. As illustrated in Fig.~\ref{fig:concept}, we assume such object property can be latent coded and shared amount domains. Following a style transfer terminology, we call such shared property as content, denoted as $c \in \Cdm$. Also, we assume the content can be encoded from all three domains.
	
	\noindent\textbf{Assumption-2. Reflectance-shading independence.} \\ 
	Physically, reflectance is the invariance against lighting and orientation while shading is the variance. And therefore, to decompose these two components, we assume their conditional priors are independent and can be learned separately. 
	As illustrated in Fig.~\ref{fig:concept}, we denote the latent prior for reflectance as $z_R \in \Rpdm$ which can be encoded from both the reflectance domain and the natural image domains. 
	Similarly, we define the latent prior for shading, $z_S \in \Spdm$. 
	
	\noindent\textbf{Assumption-3. The latent code encoders are reversible.} \\ This assumption is widely used in image-to-image translation ~\cite{UII:Huang2018MUNIT,UII:Lee2018Diverse}. In detail, it assumes an image can be encoded into the latent code, which can be decoded to image at the same time.
	This allows us to transfer style and contents among domains. Particularly, this allow us to transfer natural image to reflectance and shading.
	
	\subsection{Implementation}
	
	\begin{figure*} [!tb]
		\begin{center}
			\includegraphics[width=0.8\textwidth]{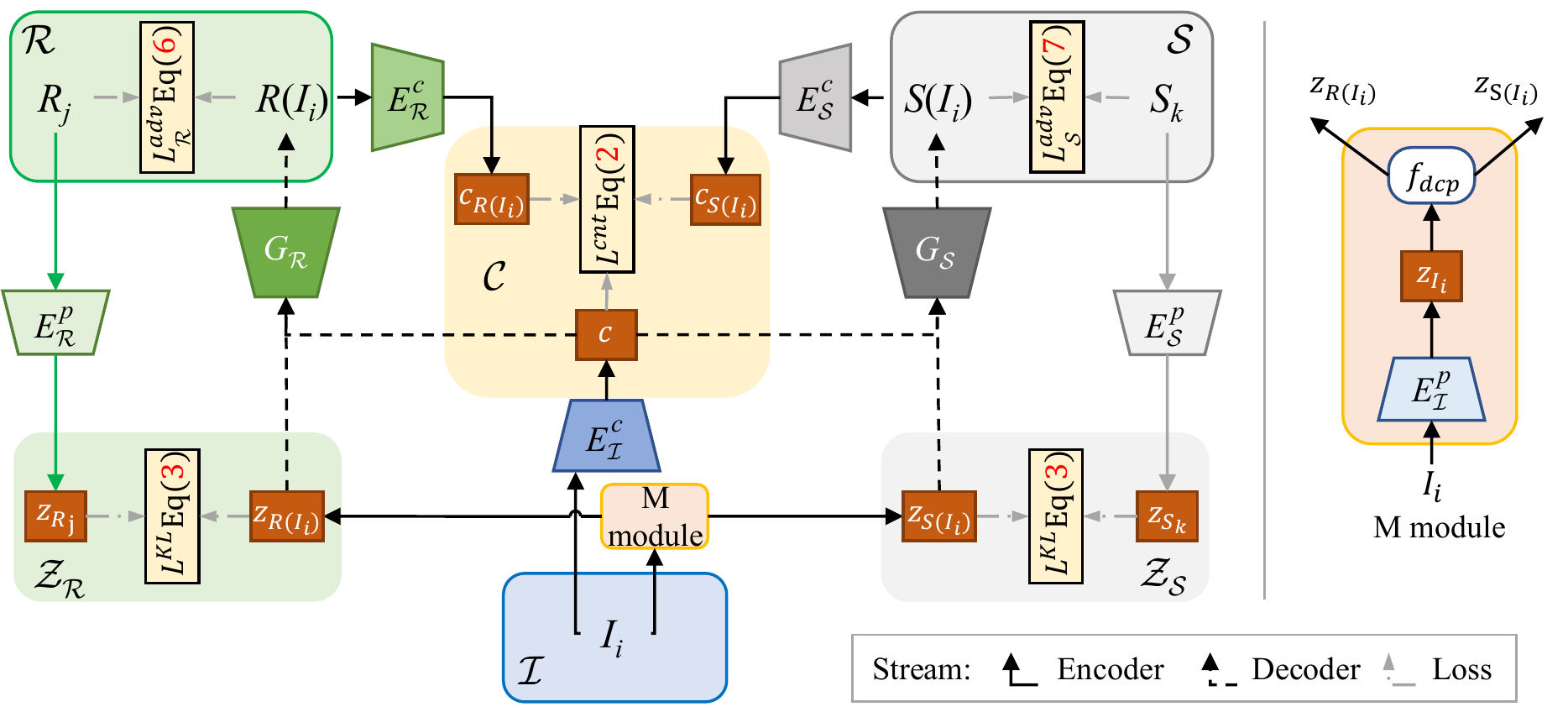}
		\end{center}
		\caption{ The proposed architecture of \proposed. Our method decompose intrinsic images with unsupervised learning manner, which translate images from natural image domain $\Idm$ to reflectance domain $\Rdm$ and shading domain $\Sdm$. 
		}
		\label{fig:main_figure}
		\vspace{-3mm}
	\end{figure*}
	
	The detailed implementation of \proposed~ network is illustrated in Fig.~\ref{fig:main_figure}. 
	
	\noindent\textbf{The Content-sharing architecture.} 
	As with Assumption-1, we design our content-sharing architecture. We use encoder $\EnIc$ to extract the content code $c$ of the input image $\Isa$, then $c$ is used to generate the decomposition layers $R(I_i)$ and $S(I_i)$ through generators $\GenR$ and $\GenS$, respectively. 
	Next, we extract the content code $c_{R(I_i)}$ of $R(I_i)$  and the content code $c_{S(I_i)}$ of $S(I_i)$ by using $\EnRc$, $\EnSc$, respectively. Finally, we apply \textit{content consistent loss} to make content encoders $\EnIc$, $\EnRc$ and $\EnSc$ work correctly. In detail, we use the content consistency $\LL^{cnt}$ to constrain the content code among input image $\Isa$ and its predictions $\Resm$ and $\Sesm$.
	\begin{equation} \label{eq:content_consistency}
	\LL^{cnt} = |c_{R(I_i)} - c |_1 + | c_{S(I_i)} - c |_1,
	\end{equation}
	where $|\centerdot|_1$ is \textit{L1} distance.
	
	\noindent\textbf{Mapping module (M module).}
	Following the Assumption-2, the prior codes of reflectance and shading are domain-variant and independent to each other. Because we need to infer the prior code $z_{R(I_i)}$ and $z_{S(I_i)}$ from $\Isa$, we design the Mapping module (M module), as shown in Fig.~\ref{fig:main_figure}. Specifically, we first extract the natural image prior code $z_{I_i}$, then we designed a decomposition mapping $f_{dcp}$ to infer the prior code $z_{R(I_i)}$ and $z_{S(I_i)}$. To constrain the $z_{R(I_i)}$ in the reflectance prior domain $\Rpdm$, we use Kllback-Leibler Divergence (KLD) and other real prior $\zR$ which is sampled from $\Rpdm$. $Z_{S(I_i)}$ is generated and constrained in the similar way. The definition \textit{KLD loss} is
	\begin{equation} \label{eq:kld_loss}
	\LL^{KL} = \E [\log p(\hat{z}) - \log q(z)],
	\end{equation}
	where the prior code $\hat{z}$ is extracted from \emph{M module} and its real prior code $z$ is extracted from its real image. Here are two prior domains $\Rpdm$ and $\Spdm$, so the total KLD loss is $\LL^{KL}_t = \E [\log p(z_{R(I_i)}) - \log q(z_{R_j})] + \E [\log p(z_{S(I_i)}) - \log q(z_{S_k})]$.
	
	\begin{figure} [!tb]
		\centering
		\includegraphics[width=\linewidth]{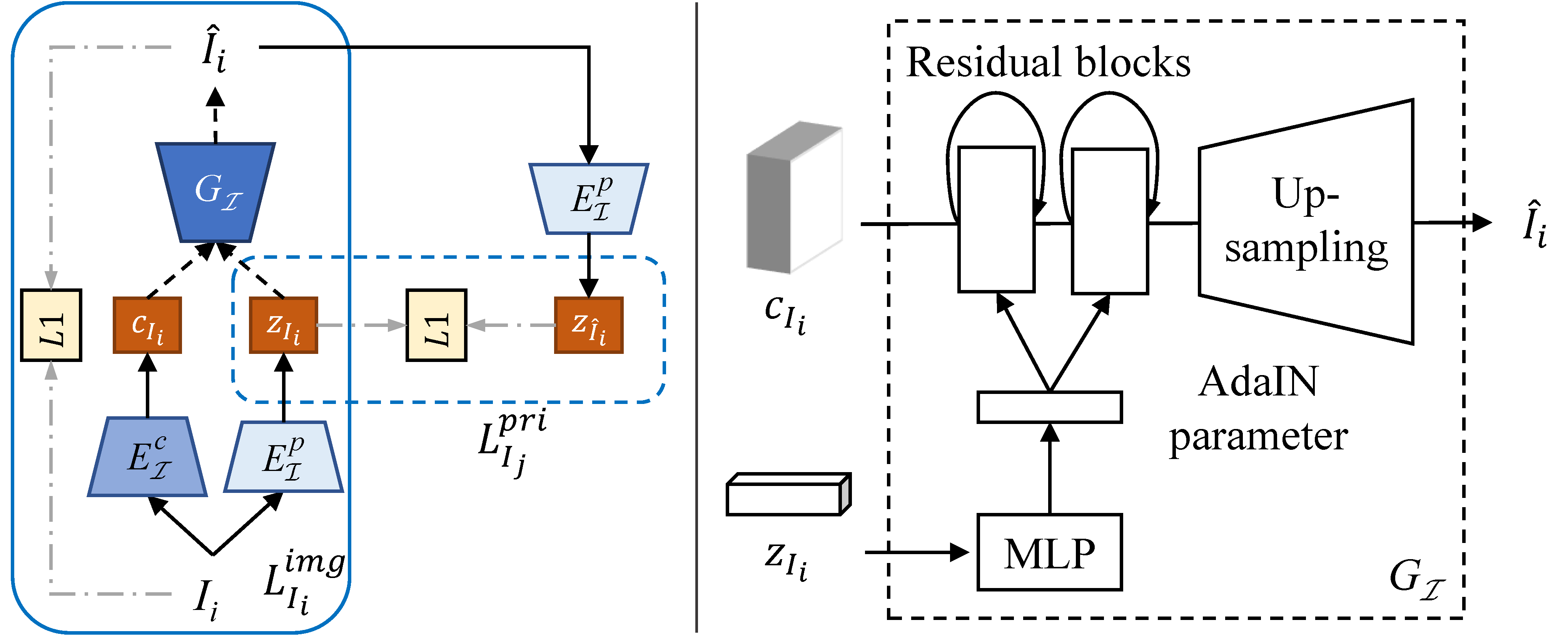}
		\caption{Left: Auto-encoder for natural image stream, the reflectance and shading streams are designed similarly. Right: Architecture of the generator. 
		}
		\label{fig:auto_encoder-generator}
		\vspace{-3mm}
	\end{figure}
	
	\noindent\textbf{Auto-encoders.}
	Per Assumption-3, we implement three auto-encoders.
	Left of Fig.~\ref{fig:auto_encoder-generator} shows the detail of implementing the auto-encoder for natural image stream. The auto-encoder for reflectance and shading are implemented in a similar way where the detail are provided in the supplementary material.
	We follow the recent image-to-image translation methods~\cite{UII:Huang2018MUNIT,UII:Lee2018Diverse} and use \textit{bidirectional reconstruction constraints} which enables reconstruction in both image $\to$ latent code $\to$ image and latent code $\to$ image $\to$ latent code directions. In detail:
	
	\emph{Image reconstruction loss.} 
	Given an image sampled from the data distribution, we can reconstruct it after encoding and decoding.
	\begin{equation} \label{eq:image_recon}
	\LL^{img} = \sum_{x\in \{\Isa\} or \{\Rsa\} or\{\Ssa\}} | G(E^c(x), E^p(x)) - x |_1.
	\end{equation}

	\begin{figure*} [htbp]
		\centering
		\includegraphics[width=\textwidth]{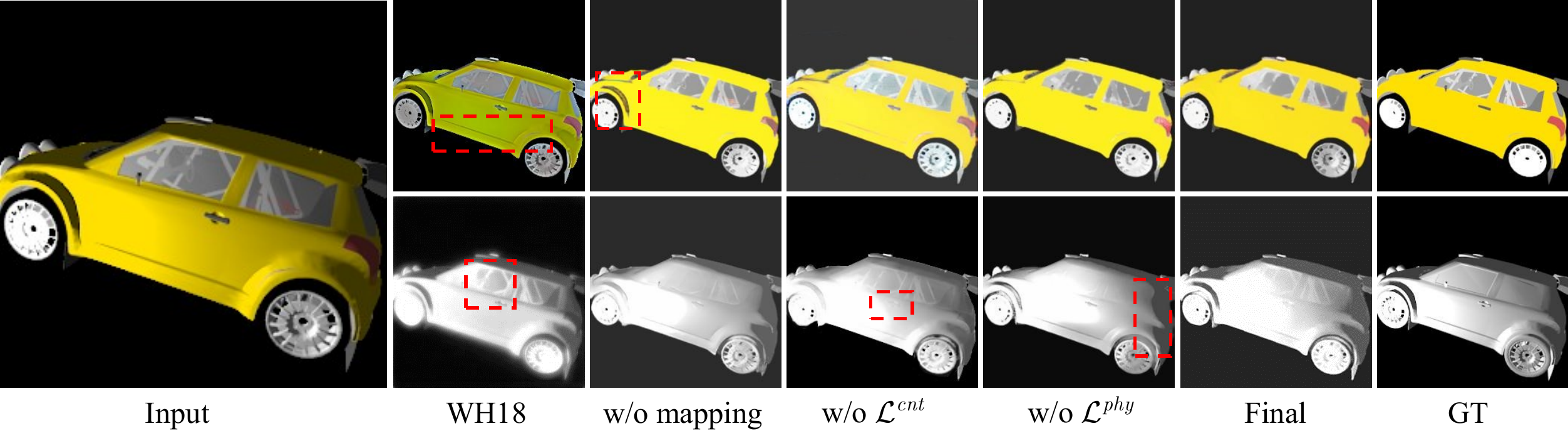}
		\caption{Visual comparison with state-of-the-art WH18~\cite{PPP:Ma2018IntrinsicDecompositionWOsingle} and ablation study on ShapeNet dataset. 
		}
		\vspace{-3mm}
		\label{fig:ablation_study}
	\end{figure*}
	
	\emph{Prior code reconstruction loss.}
	Given a prior code sampled from the latent distribution at decomposition time, we should be able to reconstruct it after decoding and encoding.
	\yf{Different from Eq.\eqref{eq:kld_loss}, which is suitable to constrain the distributions of two samples, the constraint of the prior codes of the image and the reconstructed image should be identical. Here we use $L_1$ for $\LL^{pri}$.}
	\begin{equation} \label{eq:prior_recon}
	\begin{split}
	\LL^{pri} = & |\EnIp(\GenI(c_{I_i}, z_{I_i})) - z_{I_i}|_1 \\
	+ & |\EnRp(\GenR(c_{R_j}, \zR)) - \zR|_1 \\ 
	+ & |\EnSp(\GenS(c_{S_k}, \zS)) - \zS|_1.
	\end{split}
	\end{equation}
	
	To make the decomposed intrinsic image be indistinguishable from real image in the target domain, we use GANs to match the distribution of generated images to the target data distribution. The adversarial losses are defined as follow: 
	\begin{equation} \label{eq:adv_loss_R}
	\LL^{adv}_{\Rdm} = \log (1-D_\Rdm(\Resm) + \log D_\Rdm(\Rsa),
	\end{equation}
	\begin{equation} \label{eq:adv_loss_S}
	\LL^{adv}_{\Sdm} = \log (1-D_\Sdm(\Sesm)) + \log D_\Sdm(\Ssa).
	\end{equation}
	The total adversarial loss is $\LL^{adv}_t = \LL^{adv}_{\Rdm} + \LL^{adv}_{\Sdm}$.
	
	We notice that Eq.~\eqref{eq:intro} means that image $\Isa$ is equal to the pixel-product of its analogous $\Resm$ and $\Sesm$, thus this \textit{physical loss} can be employed to regularize our method.
	\begin{equation} \label{eq:physical_loss}
	\LL^{phy} = |\Isa - \Resm\odot \Sesm|_1.
	\end{equation}
	
	\noindent\textbf{Total loss.} 
	By using the GAN scheme, we jointly train the encoders $E$, decoders $G$, mapping function $f$ and discriminators $D$ to optimize the weighted sum of the different loss terms.
	\begin{equation} \label{eq:objective_func}
	\begin{split}
	&\min_{E, G, f}\max_{D}(E, G, f, D) =  \LL^{adv}_t + \lambda_1\LL^{cnt} + \lambda_2\LL^{KL} \\ 
	& + \lambda_3\LL^{img} + \lambda_4\LL^{pri} + \lambda_5\LL^{phy},
	\end{split}
	\end{equation}
	where $\lambda_1$, $\lambda_2$, $\lambda_3$, $\lambda_4$, and $\lambda_5$ are weights that are control the importance of different loss terms.

	
	\section{Experiments}
	
	\subsection{Implementation details}
	
	\noindent\textbf{Encoder and generators.} 
	The distribution prior encoder $E^p$ is constructed by several strided convolutional layers to downsample the input image, then followed by a global average pooling layer and a dense layer.
	Our content encoder $E^c$ includes several strided convolutional layers and several residual blocks~\cite{IU:He2015DeepResidual} to downsample the input.
	All convolutional layers are followed by Instance Normalization~\cite{IU:Dmitry2015InstanceNorm}. The detailed architecture of encoders are provided in the supplementary materials.
	
	We implement the distribution prior mapping function $f_{dcp}$ via multi-layer perceptron (MLP). 
	More specifically, the $f_{dcp}$ takes $z_{I_i}$ as input, and output the concatenation of $z_{R(I_i)}$ and $z_{S(I_i)}$. 
	
	The generator reconstructs the image from its content feature and distribution prior. As illustrated in the right of Fig.~\ref{fig:auto_encoder-generator}, it processes the content feature by several unsampling and convolutional layers.
	Inspired by recent unsupervised image-to-image translation works that use affine transformation parameters in normalization layers to represent image styles~\cite{UII:Huang2017AdaIn,UII:Huang2018MUNIT}, hence we use image style to represent intrinsic image distribution priors.
	To this end, we equip the Adaptive Instance Normalization (AdaIN)~\cite{UII:Huang2017AdaIn} after each of convolutional layers in residual blocks.
	The parameters of AdaIN are dynamically generated by a MLP from the distribution prior.
	\begin{equation} \label{eq:adaIN}
	\text{AdaIN}(m, \gamma, \beta) = \gamma (\frac{m - \mu(m)}{\sigma(m)}) + \beta,
	\end{equation}
	where $m$ is the activation of the previous convolutional layer, $\mu$ and $\sigma$ are channel wise mean and standard deviation.
	$\gamma$ and $\beta$ are parameters generated by the MLP.
	
	\noindent\textbf{Discriminator.} We use multi-scale discriminators~\cite{UII:Wang2018MSdis} to guide the generators to generate high quality images in different scales including correct global structure and realistic details. We employ the LSGAN~\cite{UII:Mao2017LSGAN} as the objective.
	
	\noindent\textbf{Weights for loss terms.} In the total objective function Eq.~\eqref{eq:objective_func}, We follow \cite{UII:Huang2018MUNIT} and set $\lambda_1$, $\lambda_2$, $\lambda_3$, $\lambda_4$ as 10.0, 0.1, 10.0 and 0.1, respectively. Based on the start value and convergence of $\LL^{phy}$, we set $\lambda_5$ as 5.0 empirically.

	\subsection{Experimental setup}
	We have compared three types of intrinsic decomposition algorithms in the experiments.
	1) \textit{Unsupervised algorithms.} Retinex~\cite{PPP:Land1971retinex}, Barron \etal~\cite{PPP:Barron2015Shape}, LM14~\cite{PPP:Li2014Single}, and $\text{L}_1$ flattening that tackle image intrinsic decomposition though optimization without deep learning.
	LS18~\cite{PPP:Li2018BigTime}, WH18~\cite{PPP:Ma2018IntrinsicDecompositionWOsingle} are trained on image sequences in which each sequence include fixed reflectance and variant shadings.
	2) \textit{Image-to-image translation method.} MUNIT~\cite{UII:Huang2018MUNIT} is one of the state-of-the-art image-to-image translation methods. Because the framework of MUNIT can only translate images between two domains, so we train this method twice for input $\leftrightarrow$ reflectance and input $\leftrightarrow$ shading for generating the reflectance and shading of the input image.
	3) \textit{Supervised methods.} We have also provided results of state-of-the-art supervised methods like Zhou~\etal~\cite{PPP:Zhou2015dataDrivenIntrinsic}, MSCR~\cite{PPP:narihira2015direct_MSCR} and FY18~\cite{PPP:Fan2018revisitingDeepIntrinsic} as references.

	\subsection{Qualitative and quantitative results}
	
	We evaluate the proposed \proposed~ on four public intrinsic image benchmarks which are commonly used in image intrinsic decomposition.
	
	\begin{figure*} [htbp]
		\centering
		\includegraphics[width=\textwidth]{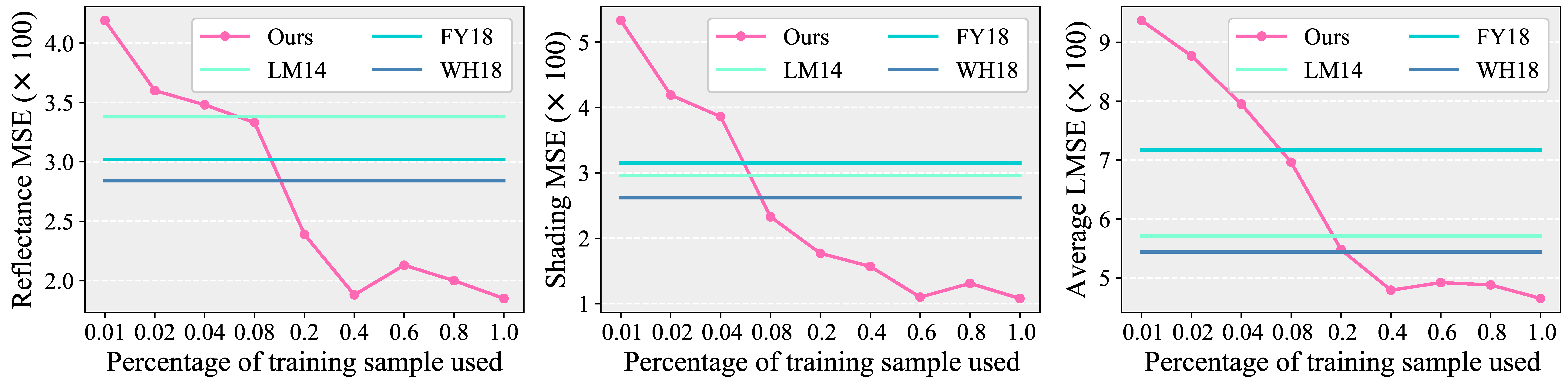}
		\caption{Performance of our unsupervised method on ShapeNet when using fewer training samples.}
		\label{fig:perform_vs_data}
	\end{figure*}

	\begin{table}[htbp]
		\begin{center}
			\setlength{\tabcolsep}{1mm}
			\caption{Numerical comparison and ablation study on ShapeNet intrinsic dataset. }
			\begin{tabular}{ccccc}
				\toprule[1.2pt]
				
				\multicolumn{ 1}{l}{} & \multicolumn{ 3}{c}{MSE} & LMSE \\
				\multicolumn{ 1}{l}{Method} &                             Reflectance &     Shading &          Avg. &       Total  \\
				\hline
				\specialrule{0em}{1pt}{1pt}
				
				LM14 \cite{PPP:Li2014Single}  	   &	0.0338  &   0.0296 &    0.0317   &  0.0623 \\
				FY18 \cite{PPP:Fan2018revisitingDeepIntrinsic}  &  0.0302 &   0.0315  & 	   0.0309 &    0.0717 \\
				WH18~\cite{PPP:Ma2018IntrinsicDecompositionWOsingle}     &  0.0284 &  0.0262  &   0.0273 &    0.0544  \\
				
				\hline
				\specialrule{0em}{1pt}{1pt}
				
				Ours w/o mapping 	       & 0.0211  &  0.0130  & 0.0171   & 0.0509   \\
				Ours w/o $\LL^{cnt}$  & 0.0239  &  0.0167  & 0.0203   & 0.0529  \\
				Ours w/o $\LL^{phy}$ 	   & 0.0196  &  0.0151  & 0.0174   & 0.0531  \\
				Ours final  & \ours{0.0185} &  \ours{0.0108}    & \ours{0.0147}  & \ours{0.0465} \\
				
				\bottomrule[1.2pt]
			\end{tabular}
		\end{center}
		\label{tab:numerical_comparisons_shapenet}
	\end{table}
	
	\begin{figure} [!tb]
		\centering
		\includegraphics[width=\linewidth]{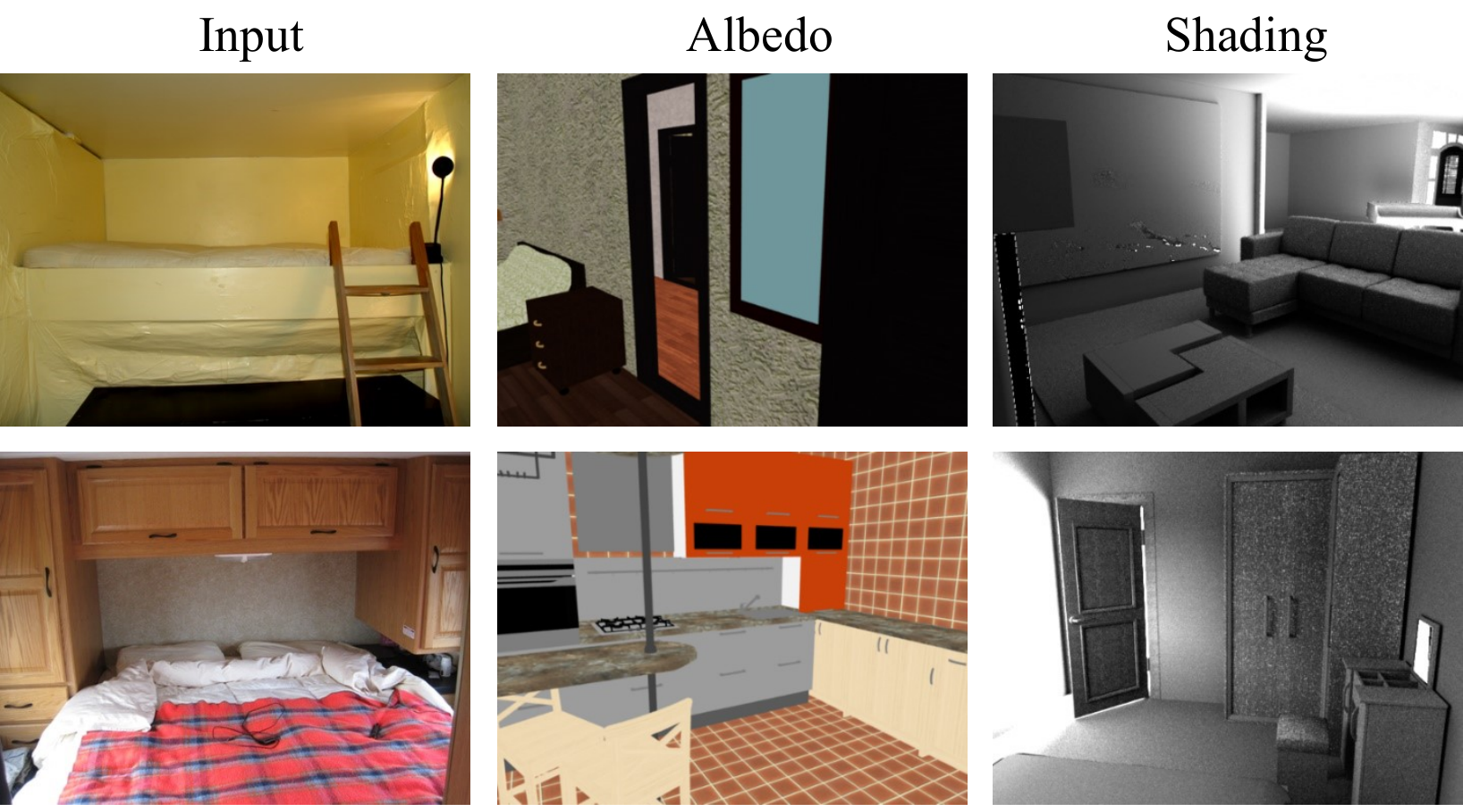}
		\caption{Samples from mixed dataset. The input images is collected from IIW, while the reflectance and shading are collected from CGIntrinsic dataset.}
		\label{fig:mixed_dataset}
		\vspace{-6mm}
	\end{figure}
	
	\noindent\textbf{ShapeNet intrinsic dataset.}
	We first make a comparison among exist methods and then perform an ablation study on this dataset.
	We use 8979 input images, 8978 reflectance images and 8978 shadings to train \proposed~, each set is shuffled before training. We use the other 2245 images for evaluation.
	
	We make a comparison with one of the optimization-based methods LM14~\cite{PPP:Li2014Single}, one of the state-of-the-art supervised methods FY18~\cite{PPP:Fan2018revisitingDeepIntrinsic} and one of the state-of-the-art unsupervised method WH18~\cite{PPP:Ma2018IntrinsicDecompositionWOsingle}. 
	 We follow settings in ~\cite{PPP:Fan2018revisitingDeepIntrinsic,PPP:narihira2015direct_MSCR} and employ mean square error (MSE), Local MSE (LMSE)~\cite{PPP:narihira2015direct_MSCR}.
	 The numerical results are listed in Table~\ref{tab:numerical_comparisons_shapenet}. The visual results are shown in Fig.~\ref{fig:ablation_study}.
	
	We employ three ablation studies to evaluate the \proposed. 
	1) The effectiveness of mapping $f_{dcp}$. $f_{dcp}$ is aiming to map the prior code of input image into reflectance prior code and shading prior code. In this ablation study, we remove it and replace the decomposed prior code with a random vector.
	2) The effectiveness of loss term $\LL^{cnt}$. $\LL^{cnt}$ is aiming to constrain the predicted reflectance and shading share the same content code with the input. We set this loss term's weight to 0 in this ablation study.
	3) The effectiveness of loss term $\LL^{phy}$. $\LL^{phy}$ is used to enforce the predicted reflectance, shading and the input have relationship as described in Eq.~\eqref{eq:intro}. We set this loss term's weight to 0 in this ablation study.
	 As shown in Fig.~\ref{fig:ablation_study}, 
	 The third column shows that without mapping, reflectance and shading aren't decomposed properly in some regions. The fourth column shows that without $\LL^{cnt}$, some details are lost in shading. The fifth column shows without $\LL^{phy}$, although shading looks fine while some area are incorrect. 
	
	To further explore how different volume of training data affect the final performance of our method, we reduce the volume of reflectance and shadings in training set and train our algorithm under the same condition. The results are illustrated in Fig.~\ref{fig:perform_vs_data}.  
	As shown in Fig.~\ref{fig:perform_vs_data}, our method can still out-perform existing methods by using as few as 20\% of training sample.
	
	\noindent\textbf{MPI-Sintel benchmark.}
	MPI-Sintel benchmark~\cite{PPP:Butler2012MPI_Sintel} is a synthesized dataset, which includes 890 images from 18 scenes with 50 frames each (except for one that contains 40 images). We follow FY18\cite{PPP:Fan2018revisitingDeepIntrinsic} and make data argumentation, after that, 8900 patches of images are generated. Then, we adopt two-fold cross validation to obtain all 890 test results.
	In the training set, we randomly select half of the input image as input samples, the reflectance and shading with remain file names as our reflectance samples and shading samples, respectively.
	
	\begin{table*}[htbp]
		\begin{center}
			\setlength{\tabcolsep}{1.3mm}
			\caption{Numerical comparison on MPI Sintel dataset. Sup. denotes that MSCR~\cite{PPP:narihira2015direct_MSCR} and FY18~\cite{PPP:Fan2018revisitingDeepIntrinsic} are fully-supervised methods, which are trained with ground truth data, and thus their results can only serve as reference.}
			\label{tab:numerical_comparisons_mpi}
			\begin{tabular}{ccccccccccc}
				\toprule[1.2pt]
				
				\multicolumn{ 2}{l}{} & \multicolumn{ 3}{c}{MSE} & \multicolumn{ 3}{c}{LMSE} & \multicolumn{ 3}{c}{DSSIM} \\
				\multicolumn{ 2}{c}{Method} &  Reflectance &     Shading &  Avg. &  Reflectance &     Shading &  Avg. &  Reflectance &     Shading &  Avg.  \\
				\hline
				\specialrule{0em}{1pt}{1pt}
				
				\multirow{2}{*}{\rotatebox{90}{\tabincell{c}{Sup.\\(Ref.)}}}
				
				& MSCR~\cite{PPP:narihira2015direct_MSCR} &  0.0100 & 0.0092 & 0.0096 & 0.0083 & 0.0085 & 0.0084 & 0.2014 & 0.1505 & 0.1760  \\
				& FY18~\cite{PPP:Fan2018revisitingDeepIntrinsic} &  0.0069 & 0.0059 & 0.0064 & 0.0044 & 0.0042 & 0.0043 & 0.1194 & 0.0822 & 0.1008  \\
				
				\hline
				\specialrule{0em}{1pt}{1pt}
				
				\multirow{5}{*}{\rotatebox{90}{Unsup.}}
				
				& Retinex~\cite{PPP:Grosse2011MITintrinsic}		 & 0.0606 & 0.0727 & 0.0667 & 0.0366 & 0.0419 & 0.0393 & 0.2270 & 0.2400 & 0.2335 \\
				& Barron \etal~\cite{PPP:Barron2015Shape}      & 0.0420 & 0.0436 & 0.0428 & 0.0298 & 0.0264 & 0.0281 & 0.2100 & 0.2060 & 0.2080 \\
				& LS18~\cite{PPP:Li2018BigTime}             & 0.0215 & 0.0251 & 0.0233 & 0.0095 & 0.0118 & 0.0107 & 0.2070 & 0.1827 & 0.1949 \\
				& MUNIT~\cite{UII:Huang2018MUNIT}              & 0.0291 & 0.0232 & 0.0262 & 0.0207 & 0.0182 & 0.0195 & 0.2073 &  0.1893 & 0.1983 \\
				& Ours         & \ours{0.0159} & \ours{0.0148} & \ours{0.0154} & \ours{0.0087} & \ours{0.0081} & \ours{0.0084} & \ours{0.1797} & \ours{0.1474} & \ours{0.1635} \\
				
				\bottomrule[1.2pt]
			\end{tabular}
		\end{center}
	\vspace{-2mm}
	\end{table*}
	
\begin{figure*} [htbp]
		\centering
		\includegraphics[width=\textwidth]{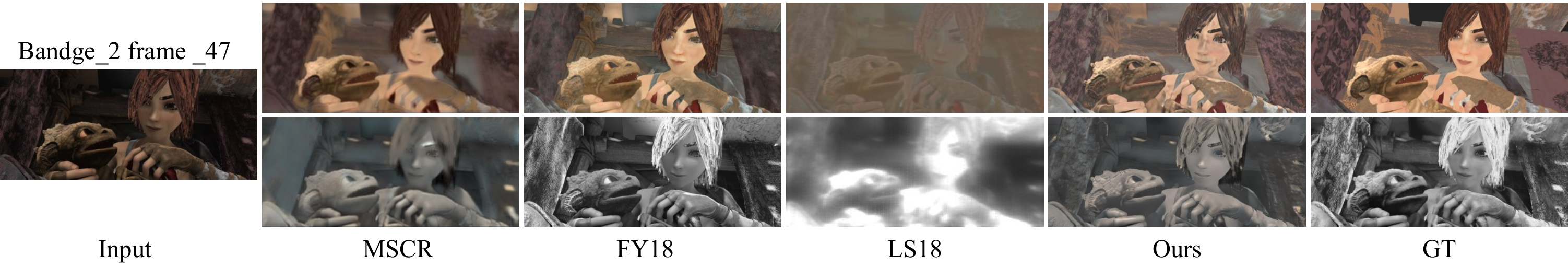}
		\caption{Visual results on MPI Sintel benchmark. Compared with state-of-the-art unsupervised method	LS18~\cite{PPP:Li2018BigTime}.	
		Supervised methods MSCR~\cite{PPP:narihira2015direct_MSCR} and FY18~\cite{PPP:Fan2018revisitingDeepIntrinsic} are provided for reference.
		}
		\label{fig:results-MPI}
		\vspace{-2mm}
	\end{figure*}

	As listed in Table~\ref{tab:numerical_comparisons_mpi},
	we employ MSE, LMSE and dissimilarity structural similarity index measure (DSSIM) as evaluation metrics. 
	We show the best of performance among unsupervised methods and even comparable with supervised method MSCR in many evaluation matrices such as LMSE and DSSIM.
	In qualitative comparison, we show one of the qualitative results in Fig.~\ref{fig:results-MPI}. The proposed \proposed~ can produce much more correct reflectance and shading than unsupervised method LS18~\cite{PPP:Li2018BigTime}. Furthermore, our results still competitive than supervised method MSCR~\cite{PPP:narihira2015direct_MSCR} since the results of MSCR with many blurry regions.
	The supervised method FY18~\cite{PPP:Fan2018revisitingDeepIntrinsic} shows the best visual results because they are trained across many other datasets and involves additional other guidance like boundary, domain filter, \etc.
	
	\begin{figure*} [!tb]
		\centering
		\includegraphics[width=\textwidth]{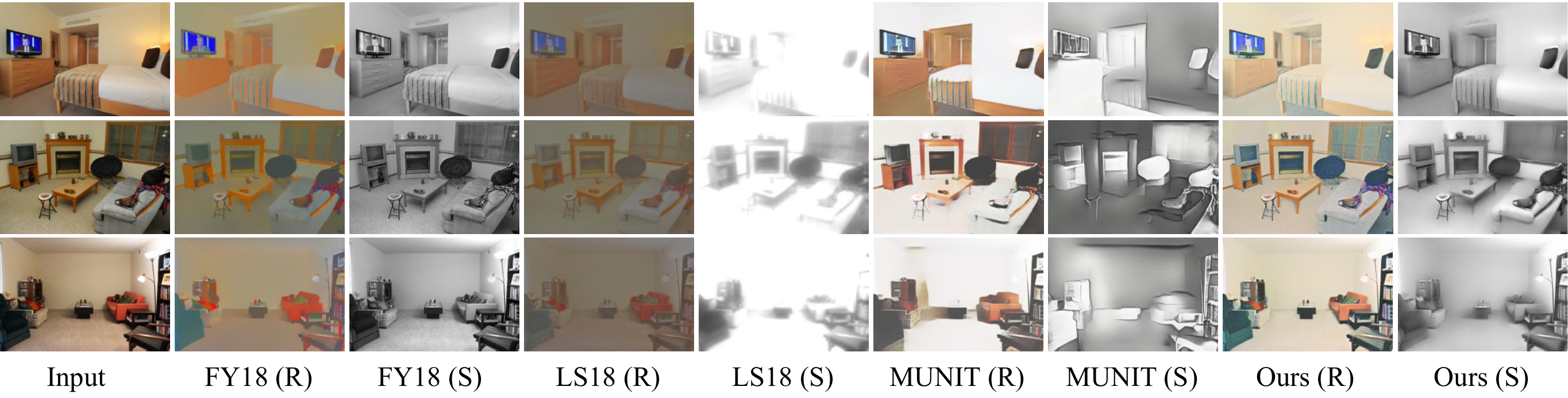}
		\caption{Qualitative comparison on the IIW test sets. 
			FY18~\cite{PPP:Fan2018revisitingDeepIntrinsic} is supervised method. LS18~\cite{PPP:Li2018BigTime} and MUNIT~\cite{UII:Huang2018MUNIT} are unsupervised.
		}
		\label{fig:iiw_results}
		\vspace{-2mm}
	\end{figure*}
	
	\noindent\textbf{MIT intrinsic dataset.}
	To test performance on real images, we use the 220 images in the MIT intrinsic dataset~\cite{PPP:Shen2011IntrinsicImages} as in \cite{PPP:narihira2015direct_MSCR}. 
	This data contains only 20 different objects, each of which has 11 images.
	To compare with previous methods, we finetune our model using 10 objects via the split from \cite{PPP:Fan2018revisitingDeepIntrinsic}, and evaluate the results using the remaining objects.
	
	\begin{table}[htbp]
		\begin{center}
			\setlength{\tabcolsep}{0.8mm}
			\caption{Numerical comparison on MIT intrinsic dataset. Results of supervised methods are also provided for reference.
			}
			\begin{tabular}{cccccc}
				\toprule[1.2pt]
				
				\multicolumn{ 2}{l}{} & \multicolumn{ 3}{c}{MSE} & LMSE \\
				\multicolumn{ 2}{c}{Method} &                             Reflectance &     Shading &          Avg. &       Total  \\
				\hline
				\specialrule{0em}{1pt}{1pt}
				
%
				\specialrule{0em}{1pt}{1pt}
				
				\multirow{4}{*}{\rotatebox{90}{\tabincell{c}{Sup. (Ref.)}}}
				
				& Zhou \etal~\cite{PPP:Zhou2015dataDrivenIntrinsic} &  0.0252 &   0.0229  & 	   0.0240 &    0.0319 \\
				& Shi \etal~\cite{PPP:Shi2017learningNonLambertian}     &  0.0216 &   0.0135  & 	   0.0175 &    0.0271 \\
				& MSCR~\cite{PPP:narihira2015direct_MSCR}            &  0.0207 &   0.0124  &        0.0165 &    0.0239  \\
				& FY18~\cite{PPP:Fan2018revisitingDeepIntrinsic}     &  \textbf{0.0134} &   \textbf{0.0089}  &  \textbf{0.0111} & \textbf{0.0203}  \\
				
				\hline
				\specialrule{0em}{1pt}{1pt}
				
				\multirow{4}{*}{\rotatebox{90}{\tabincell{c}{Unsup.}}}
				
				& LM14~\cite{PPP:Li2014Single}                   	   &	 0.0286 &     0.0227     &    0.0255    &     0.0366   \\
				& WH18~\cite{PPP:Ma2018IntrinsicDecompositionWOsingle}&    0.0232 &     0.0166     &    0.0197    &   0.0379     \\
				& MUNIT~\cite{UII:Huang2018MUNIT}                    &    0.0197 &     0.0170     &    0.0184    &   0.0302     \\
				& Ours  & \ours{0.0157} &  \ours{0.0135}    & \ours{0.0146} & \ours{0.0231} \\
				
				\bottomrule[1.2pt]
			\end{tabular}
		\end{center}
		\label{tab:numerical_comparisons_mit}
	\end{table}
	
	\begin{figure} [!tb]
		\centering
		\includegraphics[width=\linewidth]{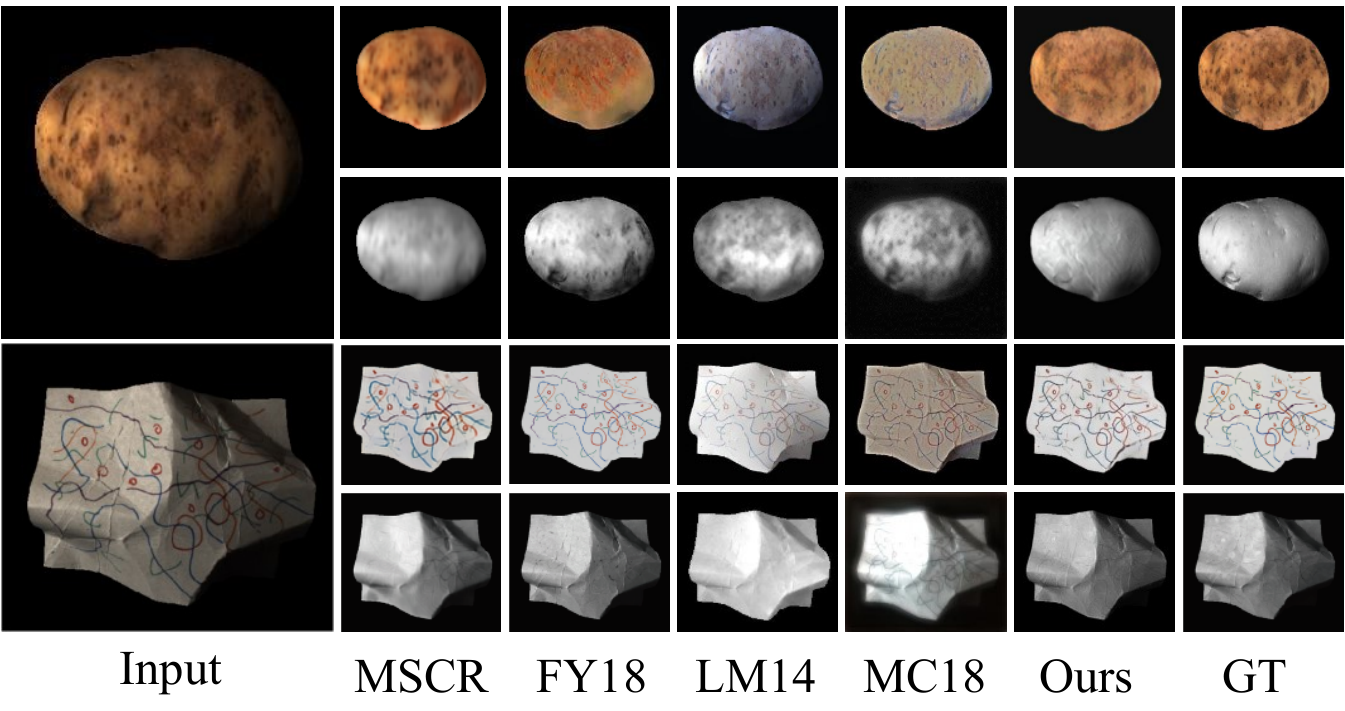}
		\caption{Visual results on MIT intrinsic image benchmark. The column 2 - 3  are supervised methods' results for reference.
		}
		\label{fig:results-MIT}
		\vspace{-7mm}
	\end{figure}

	As shown in Table~\ref{tab:numerical_comparisons_mit}, our method gets the best performance among unsupervised methods and provides comparable results among supervised methods. 
	We show two visual comparison samples in Fig.~\ref{fig:results-MIT}. Compared to the unsupervised methods, we show the best visual performance both in reflectance and shading.
	
	\noindent\textbf{IIW benchmark.} 
	The Intrinsic Images in the Wild (IIW) benchmark~\cite{PPP:bell14IIW} contains 5,230 real images of mostly indoor scenes, combined with a total of 872,161 human judgments regarding the relative reflectance between pairs of points sparsely selected throughout the images.
	We split the input images of IIW into a training set (4184) and test set in the same way as~\cite{PPP:Fan2018revisitingDeepIntrinsic,PPP:li2018cgintrinsics}.
	Because the IIW dataset contains no reflectance and shading images, we employ unpaired reflectance and shading from the rendered dataset CGIntrinsics~\cite{PPP:li2018cgintrinsics}.
	To make a fair training, we choose the first 4000 consecutive reflectance sorted by the image ID and the last 4000 shading images for training.
	Samples of the training set are illustrated in Fig.~\ref{fig:mixed_dataset}.

	\begin{table}[htbp]
		\begin{center}
			\setlength{\tabcolsep}{2.7mm}
			\caption{Numerical comparison on IIW benchmark dataset using weighted human disagreement rate (WHDR) \cite{PPP:Narihira2015WHDR}. Lower is better.
			}
			\begin{tabular}{ccc}
				\toprule[1.2pt]
				
				\multicolumn{ 2}{c}{Method} & WHDR\%(mean) \\
				\hline
				\specialrule{0em}{1pt}{1pt}
				& Baseline (const reflectance) 		& 36.54  \\
				& Baseline (const shading)         	& 51.37  \\
				
				\hline
				\specialrule{0em}{1pt}{1pt}
				
				\multirow{4}{*}{\rotatebox{90}{\tabincell{c}{Sup. (Ref.)}}}
				
				& Narihira \etal~\cite{PPP:Narihira2015WHDR}			   &  18.10  \\
				& Zhou \etal~\cite{PPP:Zhou2015dataDrivenIntrinsic}      &  15.70  \\
				& CGIntrinsic~\cite{PPP:li2018cgintrinsics}              &  14.80  \\
				& FY18~\cite{PPP:Fan2018revisitingDeepIntrinsic} 		   &  \textbf{14.45} \\
				
				\hline
				\specialrule{0em}{1pt}{1pt}
				
				\multirow{8}{*}{\rotatebox{90}{\tabincell{c}{Unsup.}}}
				
				& Retinex (color)~\cite{PPP:Grosse2011MITintrinsic} 				&  26.89    \\
				& Retinex (gray)~\cite{PPP:Grosse2011MITintrinsic} 					&  26.84    \\
				& WH18~\cite{PPP:Ma2018IntrinsicDecompositionWOsingle}     		&  28.04    \\
				& MUNIT~\cite{UII:Huang2018MUNIT}									&  25.23  \\
				& L$_1$ flattening~\cite{PPP:Bi2015IntrinsicDecompositionTOG}     &  20.94  \\
				& Bell \etal~\cite{PPP:bell14IIW} 		                        &  20.64  \\
				& LS18~\cite{PPP:Li2018BigTime}    		                        &  20.30  \\
				& Ours  												            & \ours{18.69} \\
				
				\bottomrule[1.2pt]
			\end{tabular}
		\end{center}
		\label{tab:numerical_comparisons_iiw}
	\end{table}
	
	For domain adjustment, 
	we follow CGIntrinsics and apply reflectance smoothness $\LL_{R}^{smooth}$ term to encourage reflectance predictions to be piece-wise constant.
	\begin{equation} \label{eq:refl_smooth}
	\LL_{R}^{smooth} = \sum_{i=1}^{N}\sum_{j\in \mathcal{N}(i)} v_{i,j} |\log x_R^i - \log x_R^j|_1,
	\end{equation}
	where $\mathcal{N}(i)$ denotes the 8-connected neighbourhood of the pixel at position $i$. The reflectance weight $v_{i,j} = \exp (-\frac{1}{2}(\textbf{f}_i - \textbf{f}_j)^T\Sigma^{-1}(\textbf{f}_i - \textbf{f}_j))$, and the feature vector $\textbf{f}_i$ is defined as $[\textbf{p}_i, \xx_i, r_i^1, r_i^2]$, where $\textbf{p}_i$ and $\xx_i$ are the spatial position and image intensity respectively, and $r_i^1$ and $r_i^2$ are the first two elements of chromaticity. $\Sigma$ is a covariance matrix which defines the distance between two feature vectors.
	We set this loss term with weight of 1.0 in this experiment.
	
	We use the same test split with compared method.
    Because there is no pixel-wise ground truth, we use the weighted human disagreement rate (WHDR) which is introduced in the dataset~\cite{PPP:bell14IIW}. 
	The numerical results are listed in Table~\ref{tab:numerical_comparisons_iiw}, lower is better.
	The qualitative results are illustrated in Fig.~\ref{fig:iiw_results}.
	Our proposed method generates better results than the existing methods.

	\section{Conclusion}
	In this paper, we propose an unsupervised single image intrinsic decomposition (\proposed) method. 
	We introduce three assumptions about distributions of different image domains, \ie, domain invariant content, reflectance-shading independence and the latent code encoders are reversible. We implemented our pipeline based on these assumptions.
	Experimental results on four intrinsic image benchmarks show that our method outperforms existing state-of-the-art unsupervised methods and even some state-of-the-art supervised methods. 
	
	{\newpage
		\small
		\bibliographystyle{ieee_fullname}
		\bibliography{egbib}
	}
	
\end{document}